\def\year{2019}\relax
\documentclass[letterpaper]{article} 
\usepackage{aaai19}  
\usepackage{times}  
\usepackage{helvet}  
\usepackage{courier}  
\usepackage{url}  
\usepackage{graphicx}  
\usepackage{multirow}
\usepackage{epsfig}
\usepackage{amsmath}
\usepackage{amssymb,amsfonts}
\usepackage{bm}
\usepackage{CJK}
\usepackage{subcaption}

\newcommand{\citet}[1]{\citeauthor{#1}~\shortcite{#1}}
\newcommand{\citep}{\cite}

\newcommand*{\affaddr}[1]{#1} 
\newcommand*{\affmark}[1][*]{\textsuperscript{#1}}
\newcommand*{\email}[1]{\texttt{#1}}

\begin{document}
%

\title{LiveBot: Generating Live Video Comments Based on Visual and Textual Contexts}
\author{Shuming Ma\affmark[1]\thanks{Joint work between Microsoft Research Asia and Peking University}, Lei Cui\affmark[2], Damai Dai\affmark[1], Furu Wei\affmark[2], Xu Sun\affmark[1]\\
\affaddr{\affmark[1]MOE Key Lab of Computational Linguistics, School of EECS, Peking University}\\
\affaddr{\affmark[2]Microsoft Research Asia}\\
\email{\{shumingma,daidamai,xusun\}@pku.edu.cn}\\
\email{\{lecu,fuwei\}@microsoft.com}
}

\maketitle
\begin{abstract}

We introduce the task of automatic live commenting. 
Live commenting, which is also called ``video barrage'', is an emerging feature on online video sites that allows real-time comments from viewers to fly across the screen like bullets or roll at the right side of the screen. 
The live comments are a mixture of opinions for the video and the chit chats with other comments.
Automatic live commenting requires AI agents to comprehend the videos and interact with human viewers who also make the comments, so it is a good testbed of an AI agent's ability to deal with both dynamic vision and language.  
In this work, we construct a large-scale live comment dataset with 2,361 videos and 895,929 live comments. 
Then, we introduce two neural models to generate live comments based on the visual and textual contexts, which achieve better performance than previous neural baselines such as the sequence-to-sequence model. 
Finally, we provide a retrieval-based evaluation protocol for automatic live commenting where the model is asked to sort a set of candidate comments based on the log-likelihood score, and evaluated on metrics such as mean-reciprocal-rank. 
Putting it all together, we demonstrate the first ``LiveBot''. The datasets and the codes can be found at \url{https://github.com/lancopku/livebot}.

\end{abstract}

\section{Introduction}

\begin{figure*}[t]
\centering
\includegraphics[width=0.8\linewidth]{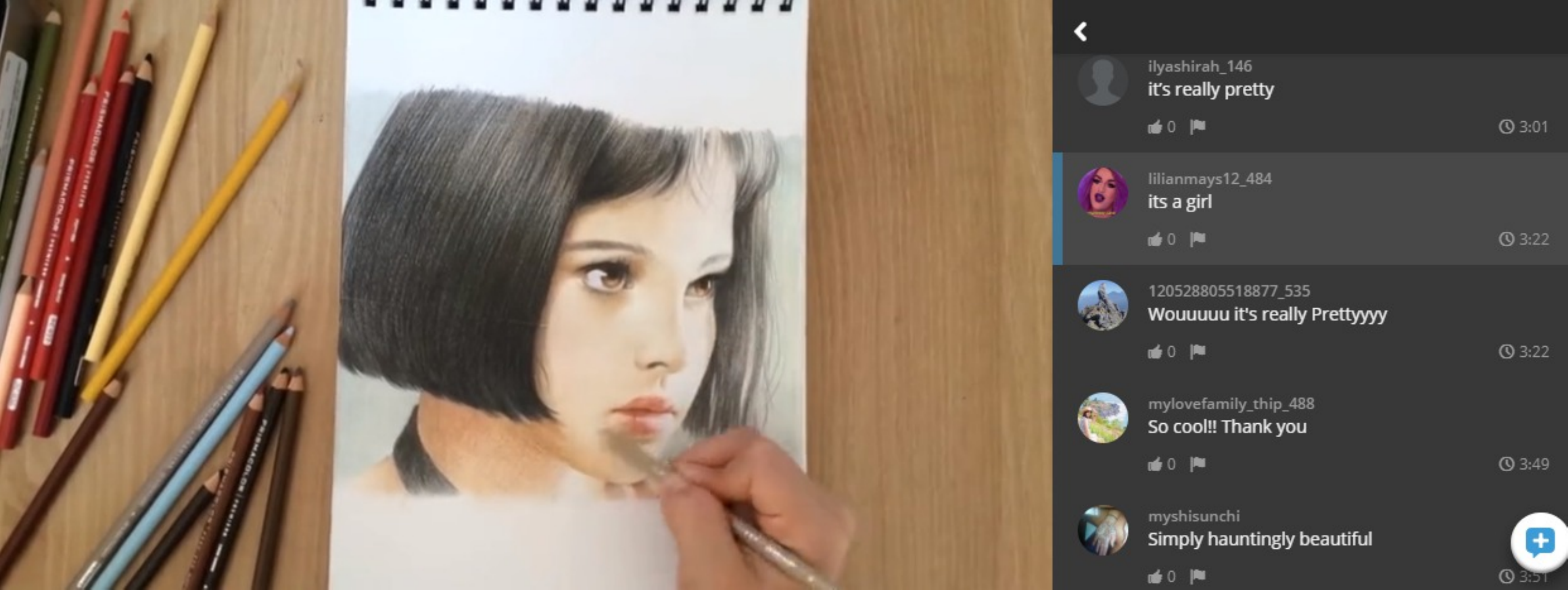}
\caption{An example of live comments from a video streaming website ViKi.}
\label{fig_example}
    \vspace{-0.1in}
\end{figure*}

The comments of videos bring many viewers fun and new ideas. Unfortunately, on many occasions, the videos and the comments are separated, forcing viewers to make a trade-off between the two key elements. To address this problem, some video sites provide a new feature: viewers can put down the comments during watching videos, and the comments will fly across the screen like bullets or roll at the right side of the screen. We show an example of live comments in Figure~\ref{fig_example}. The video is about drawing a girl, and the viewers share their watching experience and opinions with the live comments, such as ``simply hauntingly beautiful''. The live comments make the video more interesting and appealing. Besides, live comments can also better engage viewers and create a direct link among viewers, making their opinions and responses more visible than the average comments in the comment section. These features have a tremendously positive effect on the number of users, video clicks, and usage duration.

\begin{figure}[t]
	\centering
	\includegraphics[width=0.8\linewidth]{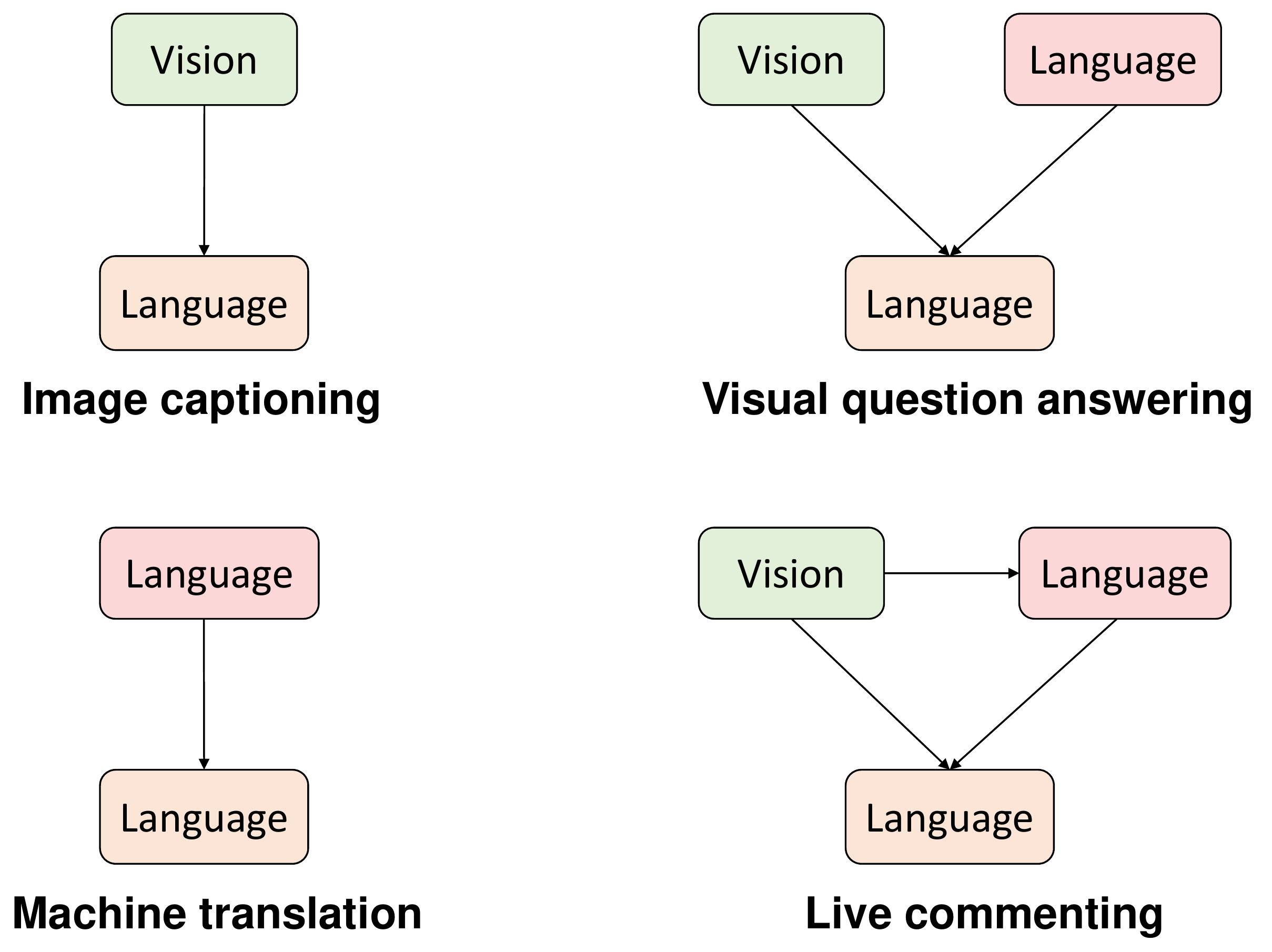}
	\caption{The relationship between vision and language in different tasks.}\label{fig_task}
    \vspace{-0.2in}
\end{figure}

Motivated by the advantages of live comments for the videos, we propose a novel task: automatic live commenting. The live comments are a mixture of the opinions for the video and the chit chats with other comments, so the task of living commenting requires AI agents to comprehend the videos and interact with human viewers who also make the comments.
Therefore, it is a good testbed of an AI agent's ability to deal with both dynamic vision and language.


With the rapid progress at the intersection of vision and language, there are some tasks to evaluate an AI's ability of dealing with both vision and language, including image captioning~\cite{ic1,ic2,ic3}, video description~\cite{vd1,vd2,vd3}, visual question answering~\cite{vqa1,vqa2}, and visual dialogue~\cite{Das2017CVPR}. Live commenting is different from all these tasks. Image captioning is to generate the textual description of an image, and video description aims to assign a description for a video. Both the two tasks only require the machine to see and understand the images or the videos, rather than communicate with the human. Visual question answering and visual dialogue take a significant step towards human-machine interaction. Given an image, the machine should answer questions about the image or conduct multiple rounds of a dialogue with the human. Different from the two tasks, live commenting requires to understand the videos and share the opinions or watching experiences, which is a more challenging task. 

A unique challenge of automatic live commenting is the complex dependency between the comments and the videos. First, the live comments are related to both the videos and the surrounding comments, and the surrounding comments also depend on the videos. We summarize the comparison of the dependency between live commenting and other tasks in Figure~\ref{fig_task}. Second, the live comments are not only conditioned on the corresponding frames that they appear on but also the surrounding frames, because the viewers may comment on either the upcoming video streaming\footnote{For example, some viewers will turn back the videos and put down the warning of upcoming surprising scenes.} or the past. More specifically, we formulate the live commenting task as: given a video $V$, one frame in the video $f$, the time-stamp $t$ of the frame, and the surrounding comments $C$ (if any) and frames $I$ at around the time-stamp, the machine should make a comment relevant to the clips or the other comments near the frame $f$.

In this work, we build a ``LiveBot'' to make live comments for the videos. 
We construct a large-scale live comment dataset with 2,361 videos and 895,929 comments from a popular Chinese video streaming website called Bilibili.
In order to model the complex dependency described above, we introduce two neural approaches to generate comments.
We also provide a retrieval-based evaluation protocol for live commenting where the model is asked to sort a set of candidate answers based on the log-likelihood score, and evaluated on metrics such as mean-reciprocal-rank. Experimental results show that our model can achieve better performance than the previous neural baselines in both automatic evaluation and human evaluation.

The contributions of the paper are as follow:
 \begin{itemize}
\item To the best of our knowledge, we are the first to propose the task of automatic live commenting for videos.
\item We construct a large-scale live comment dataset with 2,361 videos and 895,929 comments, so that the data-driven approaches are possible for this task.
\item We introduce two neural models to jointly encode the visual content and the textual content, which achieve better performance than the previous neural baselines such as the sequence-to-sequence model.
\item We provide a retrieval-based evaluation protocol for live commenting where the model is asked to sort a set of candidate answers and evaluated on metrics such as mean-reciprocal-rank. 
\end{itemize}

\begin{figure*}[t]
\centering
\subcaptionbox{0:48}{\includegraphics[width=0.32\linewidth]{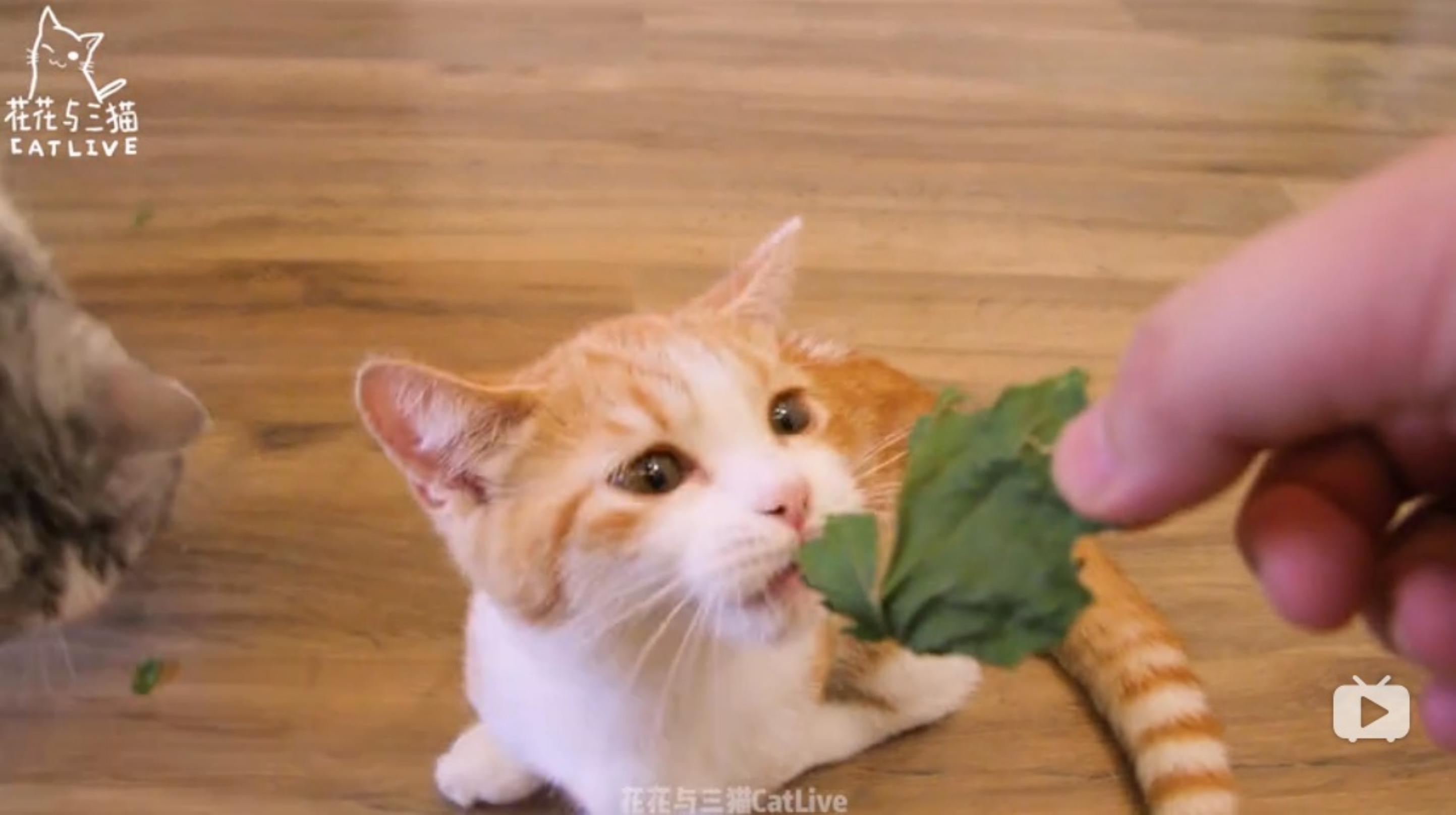}}
\subcaptionbox{1:52}{\includegraphics[width=0.32\linewidth]{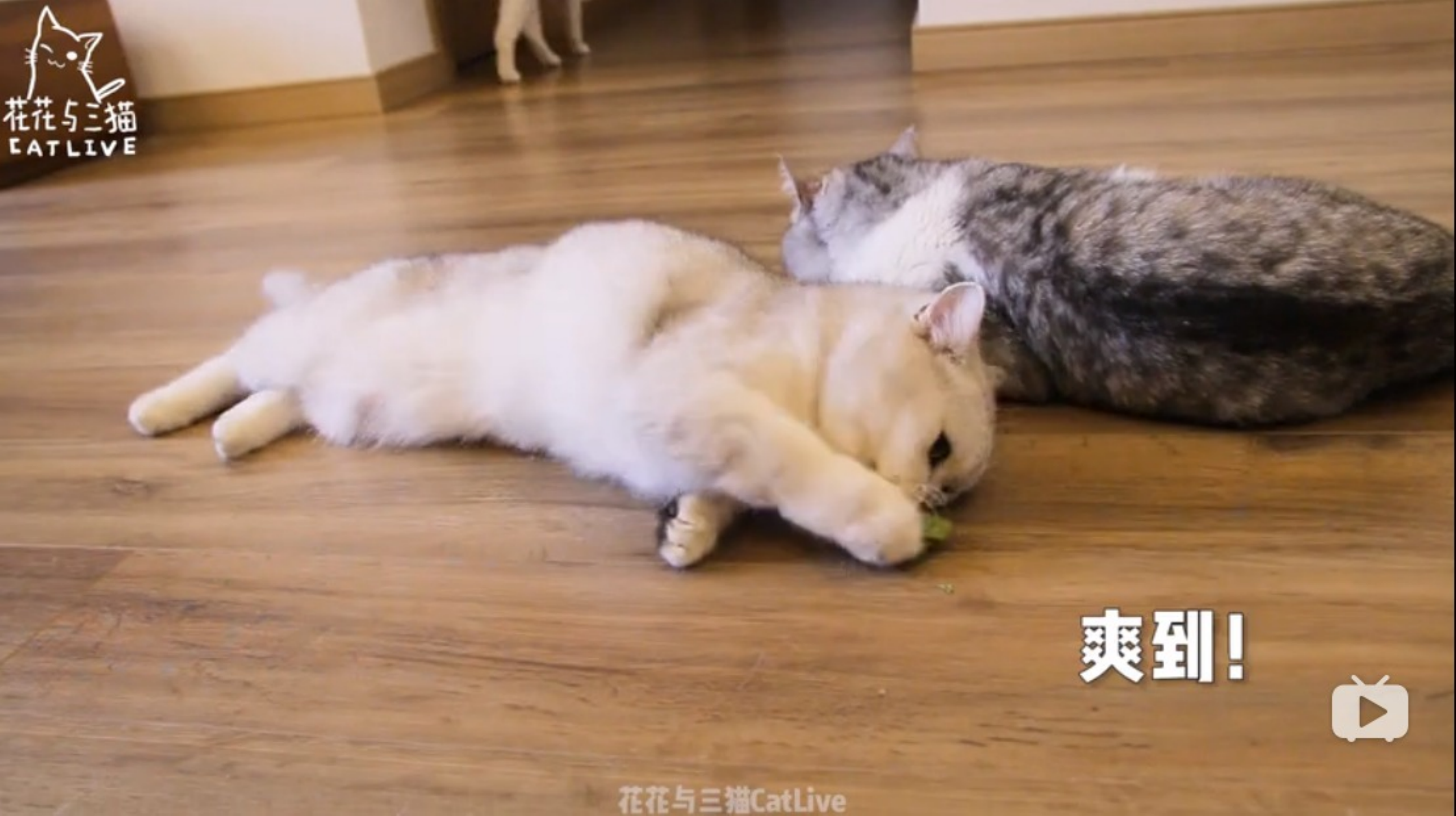}}
\subcaptionbox{3:41}{\includegraphics[width=0.32\linewidth]{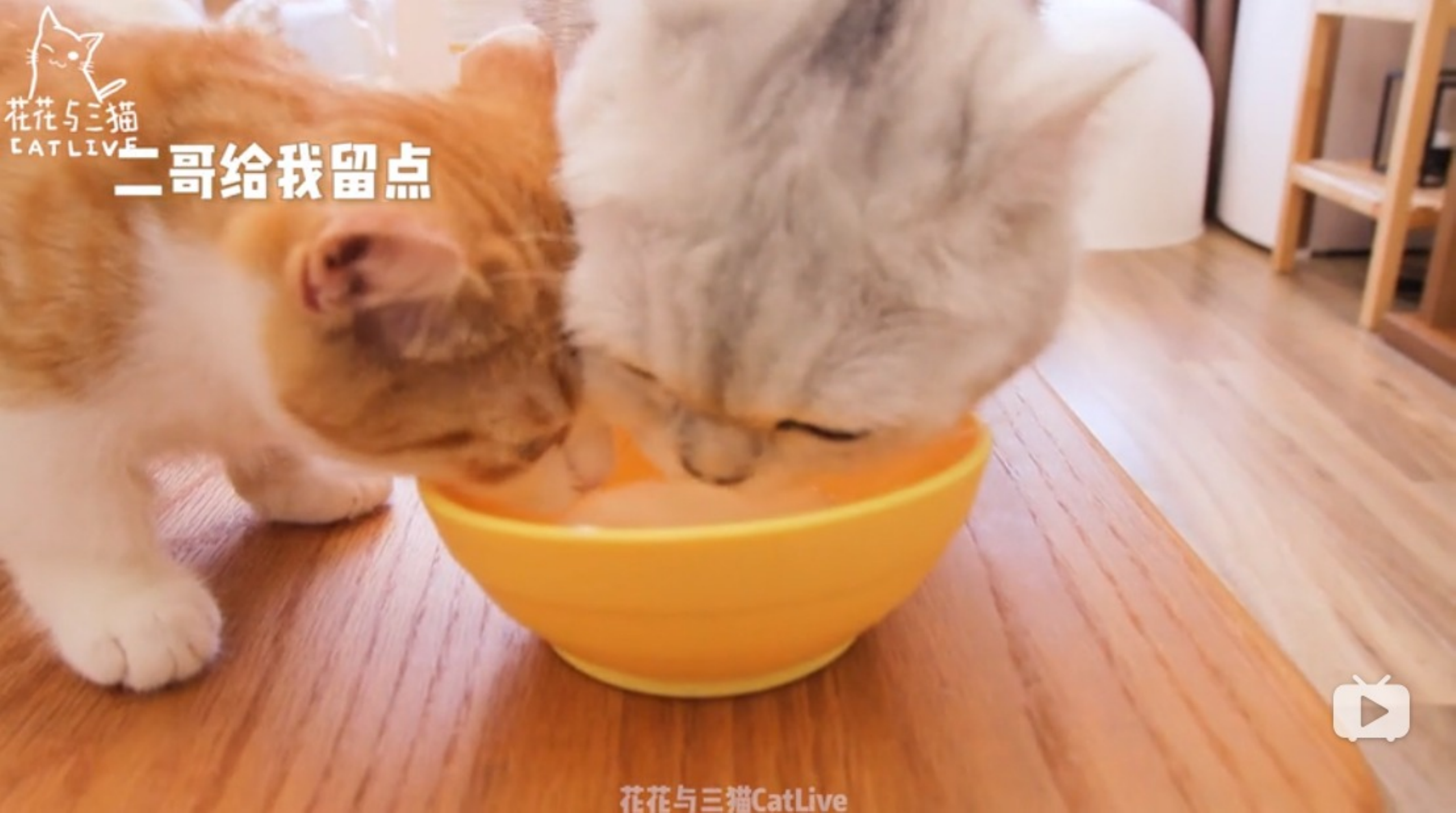}}\\
\begin{CJK}{UTF8}{gbsn}
\small
\begin{tabular}{c  p{13.5cm} }
\multicolumn{2}{c}{} \\
\hline
\multicolumn{1}{c}{Time Stamp} & \multicolumn{1}{c}{Comments} \\
\hline
0:48 & 橙猫是短腿吗 (Is the orange cat short leg?) \\
1:06 & 根本停不下来 (Simply can't stop) \\
1:09 & 哎呀好可爱啊 (Oh so cute) \\
1:52 & 天哪这么多，天堂啊 (OMG, so many kittens, what a paradise!) \\
1:56 & 这么多只猫 (So many kittens!) \\
2:39 & 我在想猫薄荷对老虎也有用吗 (I am wondering whether the catmint works for the tiger.) \\
2:41 & 猫薄荷对老虎也有用 (Catmint also works for the tiger.) \\
3:41 & 活得不如猫 (The cat lives even better than me) \\
3:43 & 两个猫头挤在一起超可爱 (It's so cute that two heads are together) \\
\hline

\end{tabular}
\end{CJK}
\caption{A data example of a video paired with selected live comments in the Live Comment Dataset. Above are three selected frames from the videos to demonstrate the content. Below is several selected live comments paired with the time stamps when the comments appear on the screen.}
\label{fig_data}
\vspace{-0.1in}
\end{figure*}

\section{The Live Comment Dataset}

In this section, we introduce our proposed Live Comment Dataset. We first describe how we collect the data and split the dataset. Then we analyze the properties of live comments. 

\subsection{Collection of Videos and Live Comments}

Here, we describe the Live Comment Dataset. The videos are collected from a popular Chinese video streaming website called Bilibili. In order to collect the representative videos, we obtain the top representative queries from the search engine, and crawl the top 10 pages of the video search results. The queries cover 19 categories, including pets, sports, animation, food, entertainment, technology and more. We remove the duplicate and short videos, and filter the videos with low quality or few live comments to maintain the data quality. As a result, we have 2,361 videos of high quality in total.

On the video website, the live comments are naturally paired with the videos. For each video, we collect all the live comments appeared in the videos. We also crawl the time-stamps when the comments appear, so that we can determine the background (surrounding frames and comments) of the given comments. We tokenize all comments with the popular python package Jieba. As a result, we have 895,929 live comments paired with the videos and the time-stamps.

We also download the audio channels of the videos. We find it intractable to align the segment of audio with the comments. Therefore, we do not segment the audio, and reserve the entire audio for the videos.

Table~\ref{fig_data} shows an example of our data. The pictures above are three selected frames to demonstrate a video about feeding cats. The table below includes several selected live comments with the corresponding time-stamps. It shows that the live comments are related to the frames where the comments appear. For example, the video describes an orange cat fed with the catmint at 0:48, while the live comment at the same frame is asking ``is the orange cat short leg?''. The comment is also related to the surrounding frames. For example, the video introduces three cats playing on the floor at 1:52, while the live comment at 1:56 is saying ``So many kittens!''. Moreover, the comment is related to the surrounding comments. For example, the comment at 2:39 asks ``whether the catmint works for the tiger'', and the comment at 2:41 responds that ``catmint also works for the tiger''.

\subsection{Dataset Split}

To split the dataset into training, development and testing sets, we separate the live comments according to the corresponding videos. The comments from the same videos will not appear solely in the training or testing set to avoid overfitting. We split the data into 2,161, 100 and 100 videos in the training, testing and development sets, respectively. Finally, the numbers of live comments are 818,905, 42,405, and 34,609 in the training, testing, development sets. Table~\ref{table-statics} presents some statistics for each part of the dataset.

\subsection{Data Statistics}

Table~\ref{table-dataset} lists the statistics and comparison among different datasets and tasks. We will release more data in the future. 
Our Bilibili dataset is among the large-scale dataset in terms of videos (2,361) and sentences (895,929).
YouCook~\cite{youcook}, TACos-M-L~\cite{tacosml} are two popular action description datasets, which focus on the cooking domain. M-VAD~\cite{mvad} and MPII-MD~\cite{vd1} are the movie description datasets, while MovieQA~\cite{movieqa} is a popular movie question answering dataset.
MSVD~\cite{msvd} is a dataset for the task of paraphrasing, and MSR-VTT~\cite{msrvtt} is for video captioning.
A major limitation for these datasets is limited domains (i.e. cooking and movie) and small size of data (in terms of videos and sentences).
Compared with these datasets, our Bilibili dataset is derived from a wide variety of video categories (19 categories), which can benefit the generalization capability of model learning.
In addition, the previous datasets are designed for the tasks of description, question answering, paraphrasing, and captioning, where the patterns between the videos and the language are clear and obvious. Our dataset is for the task of commenting, where the patterns and relationship between the videos and the language are latent and difficult to learn.
In summary, our BiliBili dataset represents one of the most comprehensive,
diverse, and complex datasets for video-to-text learning.

\begin{table}[t]
\centering
\small
\begin{tabular}{c|cccc}
\hline
\bf Statistic     & \bf Train & \bf Test & \bf Dev & \bf Total \\ \hline
\#Video    & 2,161     & 100   & 100  & 2,361     \\
\#Comment & 818,905     & 42,405    & 34,609  & 895,929   \\
\#Word     & 4,418,601    & 248,399    & 193,246  & 4,860,246  \\
Avg. Words     & 5.39    & 5.85    & 5.58  & 5.42  \\
Duration (hrs)  & 103.81     & 5.02     & 5.01  & 113.84  \\ \hline
\end{tabular}
\caption{Statistics information on the training, testing, and development sets.}
\label{table-statics}
\vspace{-0.2in}
\end{table}

\begin{table}[t]
	\centering
    \footnotesize
	\begin{tabular}{c c c c c }
		\hline
        Dataset & Task & \#Video & \#Sentence \\ 
        \hline
        YouCook & Action Description & 88 & 2,668 \\
        TACos-M-L & Action Description & 185 & 52,593 \\
        M-VAD & Movie Description & 92 & 52,593 \\
        MPII-MD & Movie Description & 94 & 68,375 \\
        MovieQA & Question Answering & 140 & 150,000 \\
        MSVD & Paraphrasing & 1,970 & 70,028 \\
        MSR-VTT & Captioning & 7,180 & 200,000 \\
        \hline
        Bilibili & Commenting & 2,361 & 895,929 \\
        \hline
    \end{tabular}
    \caption{Comparison of different video-to-text datasets.}
    \label{table-dataset}
\end{table}

\begin{table}[t]
\centering
\small
\begin{tabular}{c|ccc}
\hline
\bf Interval & \bf Edit Distance    & \bf TF-IDF & \bf Human \\ \hline
0-1s  & 11.74 & 0.048       & 4.3       \\
1-3s & 11.79 & 0.033      & 4.1     \\
3-5s  & 12.05  & 0.028   & 3.9     \\
5-10s & 12.42 & 0.025     & 3.1    \\ 
$>$10s & 12.26 & 0.015    & 2.2     \\ \hline
\end{tabular}
\caption{The average similarity between two comments at different intervals (Edit distance: lower is more similar; Tf-idf: higher is more similar; Human: higher is more similar).}
\label{table-similarity}
\vspace{-0.1in}
\end{table}

\begin{figure}[t]
\centering
\subcaptionbox{}{\includegraphics[width=0.49\linewidth]{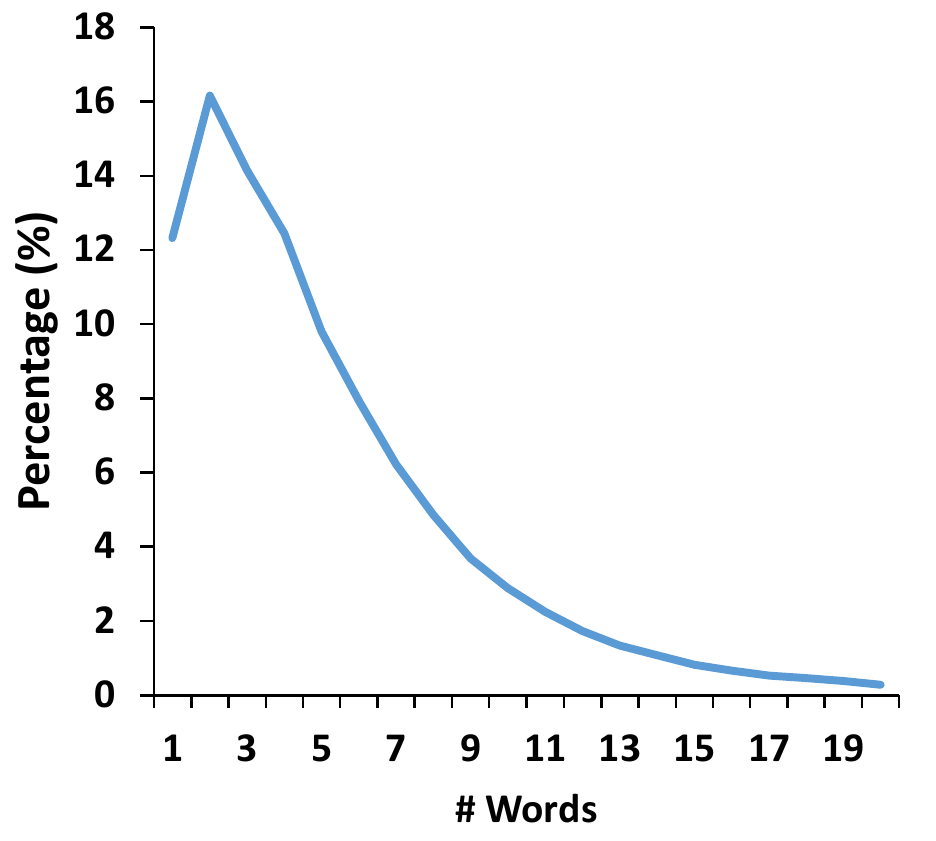}}
\subcaptionbox{}{\includegraphics[width=0.49\linewidth]{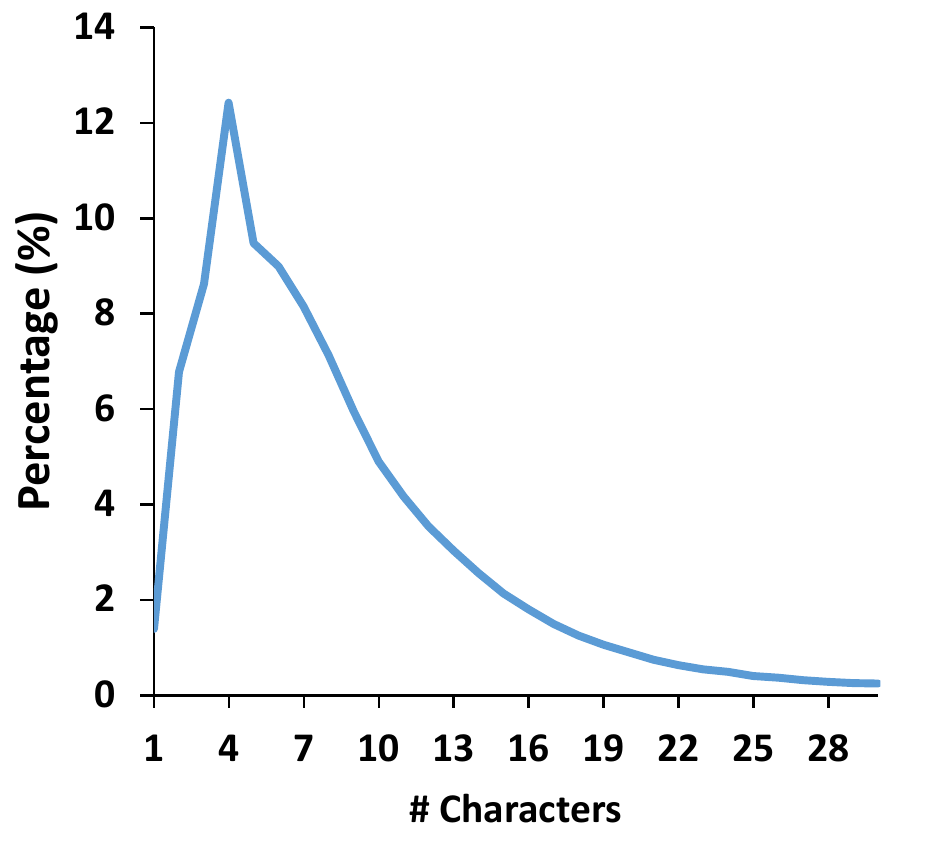}}
\caption{Distribution of lengths for comments in terms of both word-level and character-level.}
\label{fig_distribution}
\end{figure}

\subsection{Analysis of Live Comments}

Here, we analyze the live comments in our dataset. We demonstrate some properties of the live comments.

\subsubsection{Distribution of Lengths}

Figure~\ref{fig_distribution} shows the distribution of the lengths for live comments in the training set. We can see that most live comments consist of no more than 5 words or 10 characters. One reason is that 5 Chinese words or 10 Chinese characters have contained enough information to communicate with others. The other reason is that the viewers often make the live comments during watching the videos, so they prefer to make short and quick comments rather than spend much time making long and detailed comments.

\subsubsection{Correlation between Neighboring Comments}

We also validate the correlation between the neighboring comments. For each comment, we select its 20 neighboring comments to form 20 comment pairs. Then, we calculate the sentence similarities of these pairs in terms of three metrics, which are edit distance, tf-idf, and human scoring. For human scoring, we ask three annotators to score the semantic relevance between two comments, and the score ranges from 1 to 5. We group the comment pairs according to their time intervals: 0-1s, 1-3s, 3-5s, 5-10s, and more than 10s. We take the average of the scores of all comment pairs, and the results are summarized in Table~\ref{table-similarity}. It shows that the comments with a larger interval are less similar both literally and semantically. Therefore, it concludes that the neighboring comments have higher correlation than the non-neighboring ones.

\section{Approaches to Live Commenting}

A challenge of making live comments is the complex dependency between the comments and the videos. The comments are related to the surrounding comments and video clips, and the surrounding comments also rely on the videos.
To model this dependency, we introduce two approaches to generate the comments based on the visual contexts (surrounding frames) and the textual contexts (surrounding comments). 
The two approaches are based on two popular architectures for text generation: recurrent neural network (RNN) and transformer.
We denote two approaches as Fusional RNN Model and Unified Transformer Model, respectively.

\subsection{Problem Formulation}

Here, we provide the problem formulation and some notations.
Given a video $V$, a time-stamp $t$, and the surrounding comments $C$ near the time-stamp (if any), the model should generate a comment $y$ relevant to the clips or the other comments near the time-stamp. Since the video is often long and there are sometimes many comments, it is impossible to take the whole videos and all the comments as input. Therefore, we reserve the nearest $m$ frames\footnote{We set the interval between frames as 1 second.} and $n$ comments from the time-stamp $t$. We denote the $m$ frames as $\bm{I}=\{I_1,I_2,\cdots,I_m\}$, and we concatenate the $n$ comments as $\bm{C}=\{C_1,C_2,\cdots,C_n\}$. 
The model aims at generating a comment $\bm{y}=\{y_1,y_2,\cdots,y_k\}$, where $k$ is the number of words in the sentence.

\subsection{Model I: Fusional RNN Model}

\begin{figure}[t]
\centering
\includegraphics[width=1.0\linewidth]{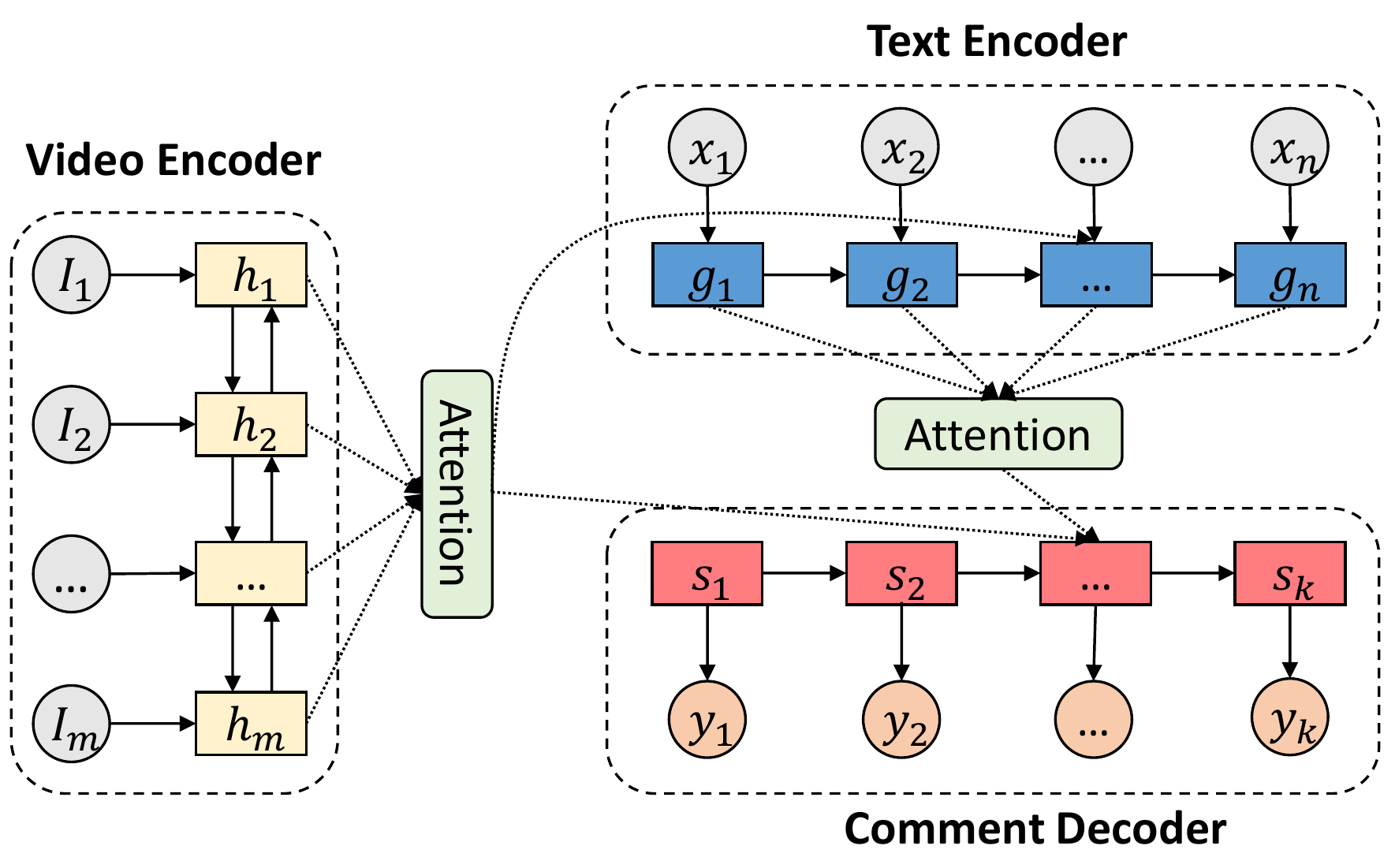}
\caption{An illustration of Fusional RNN Model.
}
\label{fig:model1}
\end{figure}

Figure~\ref{fig:model1} shows the architecture of the Fusional RNN model. The model is composed of three parts: a video encoder, a text encoder, and a comment decoder. The video encoder encodes $m$ consecutive frames with an LSTM layer on the top of the CNN layer, and the text encoder encodes $m$ surrounding live comments into the vectors with an LSTM layer. Finally, the comment decoder generates the live comment.

\subsubsection{Video Encoder}

In the video encoding part, each frame $I_{i}$ is first encoded into a vector $v_i$ by a convolution layer. We then use an LSTM layer to encode all the frame vectors into their hidden states $h_i$:
\begin{equation}
v_i = CNN(I_{i})
\end{equation}
\begin{equation}
h_i = LSTM(v_i, h_{i-1})
\end{equation}

\subsubsection{Text Encoder}

In the comment encoding part, each surrounding comment $C_i$ is first encoded into a series of word-level representations, using a word-level LSTM layer:
\begin{equation}
r_i^{j} = LSTM(C_i^j, r_i^{j-1})
\end{equation}
We use the last hidden state $r_i^{L^{(i)}}$ as the representation for $C_i$  denoted as $x_i$. Then we use a sentence-level LSTM layer with the attention mechanism to encode all the comments into sentence-level representation $g_i$:
\begin{equation}
\hat{g}_{i} = LSTM(x_i, g_{i-1})
\end{equation}
\begin{equation}
g_i=Attention(\hat{g}_{i}, \bm{h})
\end{equation}
With the help of attention, the comment representation contains the information from videos.

\subsubsection{Comment Decoder}

The model generates the comment based on both the surrounding comments and frames. Therefore, the probability of generating a sentence is defined as:
\begin{equation}
p(y_0,...,y_T|\bm{h},\bm{g}) = \prod_{t=1}^{T}p(y_t|y_0,...,y_{t-1},\bm{h},\bm{g})
\end{equation}
More specifically, the probability distribution of word $w_{i}$ is calculated as follows,
\begin{equation}
\hat{s}_i = LSTM(y_{i-1}, s_{i-1})
\end{equation}
\begin{equation}
s_i=Attention(\hat{s}_{i}, \bm{h}, \bm{g})
\end{equation}
\begin{equation}
p(w_{i}|w_0,...,w_{i-1},h) = Softmax(Ws_i)
\end{equation}

\begin{figure}[t]
\centering
\includegraphics[width=1.0\linewidth]{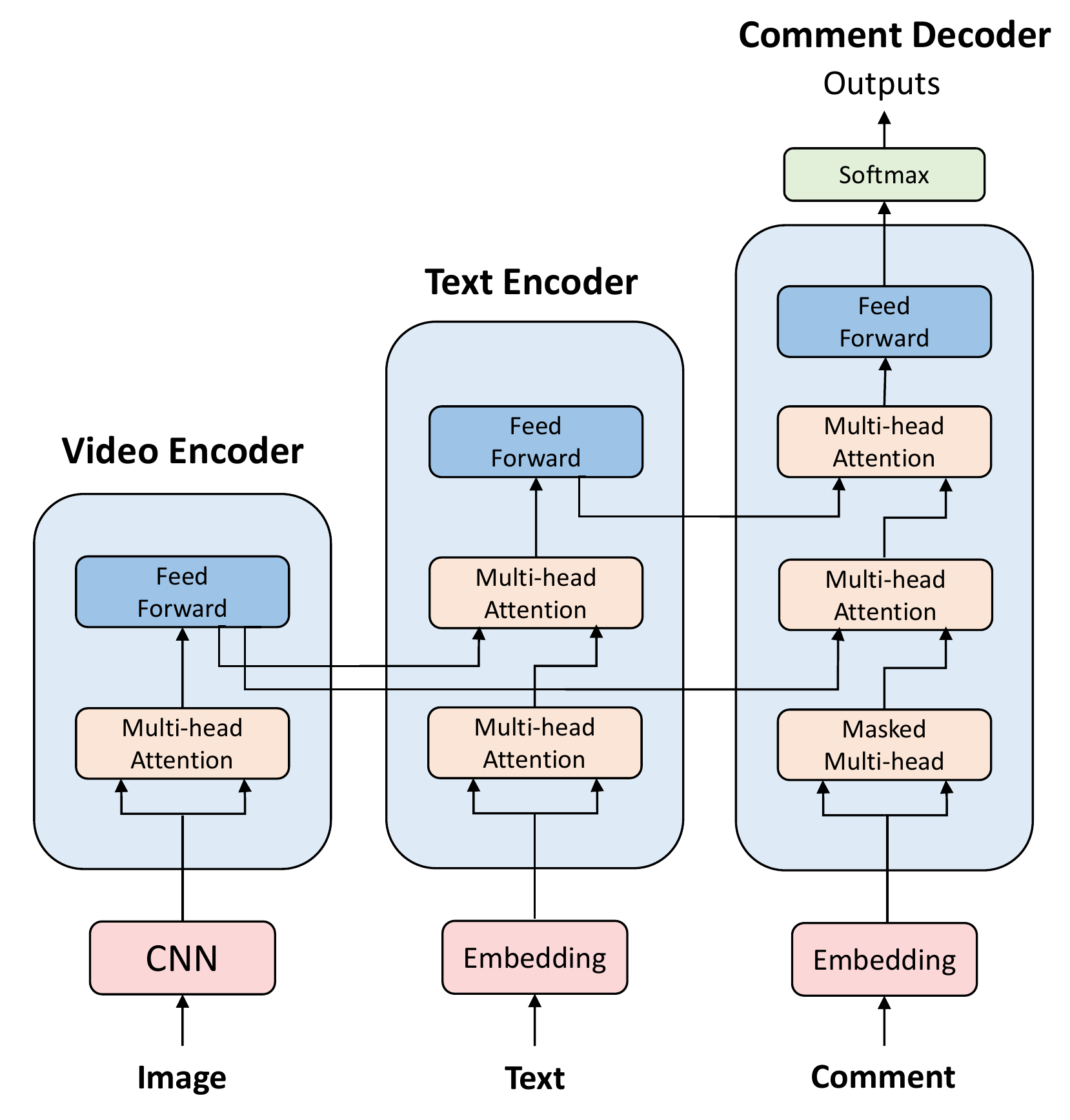}
\caption{An illustration of Unified Transformer Model.
}
\label{fig:model2}
\end{figure}

\subsection{Model II: Unified Transformer Model}

Different from the hierarchical structure of Fusional RNN model, the unified transformer model uses a linear structure to capture the dependency between the comments and the videos. Similar to Fusional RNN model, the unified transformer model consists of three parts: the video encoder, the text encoder, and the comment decoder. Figure~\ref{fig:model2} shows the architecture of the unified transformer model. In this part, we omit the details of the inner computation of the transformer block, and refer the readers to the related work~\cite{transformer}.

\subsubsection{Video Encoder}

Similar to Fusional RNN model, the video encoder first encodes each frame $I_i$ into a vector $v_i$ with a convolution layer. Then, it uses a transformer layer to encode all the frame vectors into the final representation $h_i$:
\begin{equation}
v_i = CNN(I_{i})
\end{equation}
\begin{equation}
h_i = Transformer(v_i, \bm{v})
\end{equation}
Inside the transformer, each frame's representation $v_i$ attends to a collection of the other representations $\bm{v}=\{v_1,v_2,\cdots,v_m\}$.

\subsubsection{Text Encoder}

Different from Fusion RNN model, we concatenate the comments into a word sequence $\bm{e}=\{e_1,e_2,\cdots,e_L\}$ as the input of text encoder, so that each words can contribute to the self-attention component directly. The representation of each words in the comments can be written as:
\begin{equation}
g_i = Transformer(e_i, \bm{e}, \bm{h})
\end{equation}
Inside the text encoder, there are two multi-head attention components, where the first one attends to the text input $\bm{e}$ and the second attends to the outputs $\bm{h}$ of video encoder.

\subsubsection{Comment Decoder}

We use a transformer layer to generate the live comment. The probability of generating a sentence is defined as:
\begin{equation}
p(y_0,...,y_T|\bm{h},\bm{g}) = \prod_{t=1}^{T}p(w_t|w_0,...,w_{t-1},\bm{h},\bm{g})
\end{equation}
More specifically, the probability distribution of word $y_{i}$ is calculated as:
\begin{equation}
s_i = Transformer(y_{i}, \bm{y}, \bm{h}, \bm{g})
\end{equation}
\begin{equation}
p(y_{i}|y_0,...,y_{i-1},\bm{h},\bm{g}) = Softmax(Ws_i)
\end{equation}
Inside the comment decoder, there are three multi-head attention components, where the first one attends to the comment input $\bm{y}$ and the last two attend to the outputs of video encoder $\bm{h}$ and text encoder $\bm{g}$, respectively.

\begin{table*}[t]
	\centering
	\small
	\begin{tabular}{c | l | c c |c c c c c}
		\hline
        &
		\multicolumn{1}{c|}{\textbf{Model}} &
		\multicolumn{1}{c}{\textbf{\#I}} & 
		\multicolumn{1}{c|}{\textbf{\#C}} & 
		\multicolumn{1}{c}{\textbf{Recall@1}} & 
		\multicolumn{1}{c}{\textbf{Recall@5}} &  
		\multicolumn{1}{c}{\textbf{Recall@10}} &
        \multicolumn{1}{c}{\textbf{MR}} &
        \multicolumn{1}{c}{\textbf{MRR}} \\ \hline
        \multirow{4}{*}{Video Only} &
        S2S-I & 5 & 0 & 4.69 & 19.93 & 36.46 & 21.60 & 0.1451 \\ 
        & S2S-IC & 5 & 0 & 5.49 & 20.71 & 38.35 & 20.15 & 0.1556 \\ 
        & Fusional RNN & 5 & 0 & 10.05 & 31.15 & 48.12 & 19.53 & 0.2217 \\ 
        & Unified Transformer & 5 & 0 & \textbf{11.40} & \textbf{32.62} & \textbf{50.47} & \textbf{18.12} & \textbf{0.2311} \\ 
        \hline
        \multirow{4}{*}{Comment Only}
        & S2S-C  & 0 & 5 & 9.12 & 28.05 & 44.26 & 19.76 & 0.2013 \\  
        & S2S-IC  & 0 & 5 & 10.45 & 30.91 & 46.84 & 18.06 & 0.2194 \\
        & Fusional RNN & 0 & 5 & 13.15 & \textbf{34.71} & \textbf{52.10} & 17.51 & 0.2487 \\ 
        & Unified Transformer & 0 & 5 & \textbf{13.95} & 34.57 & 51.57 & \textbf{17.01} & \textbf{0.2513} \\ 
        \hline
        \multirow{3}{*}{Both}
        & S2S-IC  & 5 & 5 & 12.89 & 33.78 & 50.29 & 17.05 & 0.2454 \\
        & Fusional RNN & 5 & 5 & 17.25 & 37.96 & \textbf{56.10} & 16.14 & 0.2710 \\ 
        & Unified Transformer & 5 & 5 & \textbf{18.01} & \textbf{38.12} & 55.78 & \textbf{16.01} & \textbf{0.2753} \\ 
        \hline
        
	\end{tabular}
	\caption{The performance of the baseline models and the proposed models. (\#I: the number of input frames used at the testing stage; \#C: the number of input comments used at the testing stage; Recall@k, MRR: higher is better; MR: lower is better)}\label{tab_res}
    \vspace{-0.15in}
\end{table*}

\begin{table}[t]
	\centering
	\small
    \setlength{\tabcolsep}{2.0mm}{
        \begin{tabular}{l|ccc}
            \hline
            \textbf{Model} & \textbf{Fluency} & \textbf{Relevance} & \textbf{Correctness} \\ 	
            \hline
            S2S-IC & 4.07 & 2.23 &  2.91 \\
            Fusion  & \textbf{4.45} & 2.95 & 3.34 \\
            Transformer & 4.31 & \textbf{3.07} & \textbf{3.45} \\
            \hline
            Human & 4.82 & 3.31 & 4.11 \\ \hline
        \end{tabular}
    }
	\caption{Results of human evaluation metrics on the test set (higher is better). All these models are trained and tested give both videos and surrounding comments. 
    }\label{manual}
    \vspace{-0.15in}
\end{table}

\section{Evaluation Metrics}

The comments can be various for a video, and it is intractable to find out all the possible references to be compared with the model outputs. Therefore, most of the reference-based metrics for generation tasks like BLEU and ROUGE are not appropriate to evaluate the comments.
Inspired by the evaluation methods of dialogue models~\cite{Das2017CVPR}, we formulate the evaluation as a ranking problem. The model is asked to sort a set of candidate comments based on the log-likelihood score. Since the model generates the comments with the highest scores, it is reasonable to discriminate a good model according to its ability to rank the correct comments on the top of the candidates. The candidate comment set consists of the following parts:
\begin{itemize}
\item\textbf{Correct:} The ground-truth comments of the corresponding videos provided by the human.

\item\textbf{Plausible}: The 50 most similar comments to the title of the video. We use the title of the video as the query to retrieval the comments that appear in the training set based on the cosine similarity of their tf-idf values. We select the top 30 comments that are not the correct comments as the plausible comments.

\item\textbf{Popular:} The 20 most popular comments from the dataset. We count the frequency of each comment in the training set, and select the 20 most frequent comments to form the popular comment set. The popular comments are the general and meaningless comments, such as ``2333'', ``Great'', ``hahahaha'', and ``Leave a comment''. These comments are dull and do not carry any information, so they are regarded as incorrect comments.

\item \textbf{Random:} After selecting the correct, plausible, and popular comments, we fill the candidate set with randomly selected comments from the training set so that there are 100 unique comments in the candidate set.
\end{itemize}

Following the previous work~\cite{Das2017CVPR}, We measure the rank in terms of the following metrics:
%
\textbf{Recall@k} (the proportion of human comments found in the top-k recommendations),
%
\textbf{Mean Rank} (the mean rank of the human comments),
%
\textbf{Mean Reciprocal Rank} (the mean reciprocal rank of the human comments).


\section{Experiments}

\subsection{Settings}

For both models, the vocabulary is limited to the 30,000 most common words in the training dataset. We use a shared embedding between encoder and decoder and set the word embedding size to 512. 
For the encoding CNN, we use a pretrained resnet with 18 layers provided by the Pytorch package. 
For both models, the batch size is 64, and the hidden dimension is 512.
We use the Adam~\cite{KingmaBa2014} optimization method to train our models. 
For the hyper-parameters of Adam optimizer, we set the learning rate $\alpha = 0.0003$, two momentum parameters $\beta_{1} = 0.9$ and $\beta_{2} = 0.999$
respectively, and $\epsilon = 1 \times 10^{-8}$.

\subsection{Baselines}

\begin{itemize}

\item \textbf{S2S-I}~\citep{vinyals2015show} applies the CNN to encode the frames, based on which the decoder generates the target live comment. This model only uses the video as the input.

\item\textbf{S2S-C}~\citep{sutskever2014sequence} applies an LSTM to make use of the surrounding comments, based on which the decoder generates the comments. This model can be regarded as the traditional sequence-to-sequence model, which only uses the surrounding comments as input.

\item\textbf{S2S-IC} is similar to ~\citep{vd2} which makes use of both the visual and textual information. In our implementation, the model has two encoders to encode the images and the comments respectively. Then, we concatenate the outputs of two encoders to decode the output comments with an LSTM decoder.

\end{itemize}

\subsection{Results}

At the training stage, we train S2S-IC, Fusional RNN, and Unified Transformer with both videos and comments. S2S-I is trained with only videos, while S2S-C is trained with only comments. At the testing stage, we evaluate these models under three settings: video only, comment only, and both video and comment. Video only means that the model only uses the images as inputs (5 nearest frames), which simulates the case when no surrounding comment is available. Comment only means that the model only takes input of the surrounding comments (5 nearest comments), which simulates the case when the video is of low quality. Both denotes the case when both the videos and the surrounding comments are available for the models (5 nearest frames and comments).

Table~\ref{tab_res} summarizes the results of the baseline models and the proposed models under three settings. It shows that our proposed models outperform the baseline models in terms of all evaluation metrics under all settings, which demonstrates the effectiveness of our proposed models. Moreover, it concludes that given both videos and comments the same models can achieve better performance than those with only videos or comments. Finally, the models with only comments as input outperform the models with only videos as input, mainly because the surrounding comments can provide more direct information for making the next comments.

\subsection{Human Evaluation}

The retrieval evaluation protocol evaluates the ability to discriminate the good comments and the bad comments. We also would like to evaluate the ability to generate human-like comments. However, the existing generative evaluation metrics, such as BLEU and ROUGE, are not reliable, because the reference comments can be various. Therefore, we conduct human evaluation to evaluate the outputs of each model.

We evaluate the generated comments in three aspects:
\textbf{Fluency} is designed to measure whether the generated live comments are fluent setting aside the relevance to videos. 
\textbf{Relevance} is designed to measure the relevance between the generated live comments and the videos. 
\textbf{Correctness} is designed to synthetically measure the confidence that the generated live comments are made by humans in the context of the video. 
For all of the above three aspects, we stipulate the score to be an integer in $\{1, 2, 3, 4, 5\}$. The higher the better. The scores are evaluated by three human annotators and finally we take the average of three raters as the final result. 

We compare our Fusional RNN model and Unified Transformer model with the strong baseline model S2S-IC. All these models are trained and tested give both videos and surrounding comments.
As shown in Table~\ref{manual}, our models achieve higher scores over the baseline model in all three degrees, which demonstrates the effectiveness of the proposed models.
We also evaluate the reference comments in the test set, which are generated by the human. It shows that the comments from human achieve high fluency and correctness scores. However, the relevance score is lower than the fluency and correctness, mainly because the comments are not always relevant to the videos, but with the surrounding comments.
According to the table, it also concludes that the comments from unified transformer are almost near to those of real-world live comments. We use Spearman's Rank correlation coefficients to evaluate the agreement among the raters. The coefficients between any two raters are all near 0.6 and at an average of 0.63. These high coefficients show that our human evaluation scores are consistent and credible.


\section{Related Work}

Inspired by the great success achieved by the sequence-to-sequence learning framework in machine translation \citep{sutskever2014sequence,cho2014learning,bahdanau2014neural}, \citet{vinyals2015show} and \citet{mao2014explain} proposed to use a deep convolutional neural network to encode the image and a recurrent neural network to generate the image captions. \citet{xu2015show} further proposed to apply attention mechanism to focus on certain parts of the image when decoding. Using CNN to encode the image while using RNN to decode the description is natural and effective when generating textual descriptions. 

One task that is similar to live comment generation is image caption generation, which is an area that has been studied for a long time. \citet{farhadi2010every} tried to generate descriptions of an image by retrieving from a big sentence pool. \citet{kulkarni2011baby} proposed to generate descriptions based on the parsing result of the image with a simple language model. These systems are often applied in a pipeline fashion, and the generated description is not creative. More recent work is to use stepwise merging network to improve the performance \cite{liu2018}.

Another task which is similar to this work is video caption generation. \citet{vd2} proposed to use CNN to extract image features, and use LSTM to encode them and decode a sentence. Similar models\citep{shetty2016frame,jin2016describing,ramanishka2016multimodal,dong2016early,pasunuru2017multi,shen2017weakly} are also proposed to handle the task of video caption generation. \citet{Das2017CVPR} introduce the task of Visual Dialog, which requires an AI agent to answer a question about an image when given the image and a dialogue history. Moreover, we are also inspired by the recent related work of natural language generation models with the text inputs \cite{DBLP:journals/corr/abs-1803-01465,DBLP:journals/corr/abs-1805-05181}. 


\section{Conclusions}

We propose the tasks of automatic live commenting, and construct a large-scale live comment dataset.
We also introduce two neural models to generate the comments which jointly encode the visual contexts and textual contexts.  
Experimental results show that our models can achieve better performance than the previous neural baselines.

\section*{Acknowledgement}

We thank the anonymous reviewers for their thoughtful comments. This work was supported in part by National Natural Science Foundation of China (No. 61673028). Xu Sun is the corresponding
author of this paper.

\bibliographystyle{aaai}
\bibliography{aaai19}

\begin{thebibliography}{}

\bibitem[\protect\citeauthoryear{Agrawal, Batra, and Parikh}{2016}]{vqa1}
Agrawal, A.; Batra, D.; and Parikh, D.
\newblock 2016.
\newblock Analyzing the behavior of visual question answering models.
\newblock In {\em {EMNLP} 2016},  1955--1960.

\bibitem[\protect\citeauthoryear{Antol \bgroup et al\mbox.\egroup
  }{2015}]{vqa2}
Antol, S.; Agrawal, A.; Lu, J.; Mitchell, M.; Batra, D.; Zitnick, C.~L.; and
  Parikh, D.
\newblock 2015.
\newblock {VQA:} visual question answering.
\newblock In {\em {ICCV} 2015},  2425--2433.

\bibitem[\protect\citeauthoryear{Bahdanau, Cho, and
  Bengio}{2014}]{bahdanau2014neural}
Bahdanau, D.; Cho, K.; and Bengio, Y.
\newblock 2014.
\newblock Neural machine translation by jointly learning to align and
  translate.
\newblock {\em arXiv preprint arXiv:1409.0473}.

\bibitem[\protect\citeauthoryear{Chen and Dolan}{2011}]{msvd}
Chen, D., and Dolan, W.~B.
\newblock 2011.
\newblock Collecting highly parallel data for paraphrase evaluation.
\newblock In {\em ACL 2011},  190--200.

\bibitem[\protect\citeauthoryear{Cho \bgroup et al\mbox.\egroup
  }{2014}]{cho2014learning}
Cho, K.; Van~Merri{\"e}nboer, B.; Gulcehre, C.; Bahdanau, D.; Bougares, F.;
  Schwenk, H.; and Bengio, Y.
\newblock 2014.
\newblock Learning phrase representations using rnn encoder-decoder for
  statistical machine translation.
\newblock {\em arXiv preprint arXiv:1406.1078}.

\bibitem[\protect\citeauthoryear{Das \bgroup et al\mbox.\egroup
  }{2013}]{youcook}
Das, P.; Xu, C.; Doell, R.~F.; and Corso, J.~J.
\newblock 2013.
\newblock A thousand frames in just a few words: Lingual description of videos
  through latent topics and sparse object stitching.
\newblock In {\em {CVPR} 2013},  2634--2641.

\bibitem[\protect\citeauthoryear{Das \bgroup et al\mbox.\egroup
  }{2017}]{Das2017CVPR}
Das, A.; Kottur, S.; Gupta, K.; Singh, A.; Yadav, D.; Moura, J.~M.; Parikh, D.;
  and Batra, D.
\newblock 2017.
\newblock {Visual dialog}.
\newblock In {\em CVPR 2017},  1080--1089.

\bibitem[\protect\citeauthoryear{Donahue \bgroup et al\mbox.\egroup
  }{2017}]{ic1}
Donahue, J.; Hendricks, L.~A.; Rohrbach, M.; Venugopalan, S.; Guadarrama, S.;
  Saenko, K.; and Darrell, T.
\newblock 2017.
\newblock Long-term recurrent convolutional networks for visual recognition and
  description.
\newblock {\em {IEEE} Trans. Pattern Anal. Mach. Intell.} 39(4):677--691.

\bibitem[\protect\citeauthoryear{Dong \bgroup et al\mbox.\egroup
  }{2016}]{dong2016early}
Dong, J.; Li, X.; Lan, W.; Huo, Y.; and Snoek, C.~G.
\newblock 2016.
\newblock Early embedding and late reranking for video captioning.
\newblock In {\em Proceedings of the 2016 ACM on Multimedia Conference},
  1082--1086.
\newblock ACM.

\bibitem[\protect\citeauthoryear{Fang \bgroup et al\mbox.\egroup }{2015}]{ic2}
Fang, H.; Gupta, S.; Iandola, F.~N.; Srivastava, R.~K.; Deng, L.; Doll{\'{a}}r,
  P.; Gao, J.; He, X.; Mitchell, M.; Platt, J.~C.; Zitnick, C.~L.; and Zweig,
  G.
\newblock 2015.
\newblock From captions to visual concepts and back.
\newblock In {\em {CVPR} 2015},  1473--1482.

\bibitem[\protect\citeauthoryear{Farhadi \bgroup et al\mbox.\egroup
  }{2010}]{farhadi2010every}
Farhadi, A.; Hejrati, M.; Sadeghi, M.~A.; Young, P.; Rashtchian, C.;
  Hockenmaier, J.; and Forsyth, D.~A.
\newblock 2010.
\newblock Every picture tells a story: generating sentences from images.
\newblock  15--29.

\bibitem[\protect\citeauthoryear{Jin \bgroup et al\mbox.\egroup
  }{2016}]{jin2016describing}
Jin, Q.; Chen, J.; Chen, S.; Xiong, Y.; and Hauptmann, A.
\newblock 2016.
\newblock Describing videos using multi-modal fusion.
\newblock In {\em Proceedings of the 2016 ACM on Multimedia Conference},
  1087--1091.
\newblock ACM.

\bibitem[\protect\citeauthoryear{Karpathy and Fei{-}Fei}{2017}]{ic3}
Karpathy, A., and Fei{-}Fei, L.
\newblock 2017.
\newblock Deep visual-semantic alignments for generating image descriptions.
\newblock {\em {IEEE} Trans. Pattern Anal. Mach. Intell.} 39(4):664--676.

\bibitem[\protect\citeauthoryear{Kingma and Ba}{2014}]{KingmaBa2014}
Kingma, D.~P., and Ba, J.
\newblock 2014.
\newblock Adam: {A} method for stochastic optimization.
\newblock {\em CoRR} abs/1412.6980.

\bibitem[\protect\citeauthoryear{Kulkarni \bgroup et al\mbox.\egroup
  }{2011}]{kulkarni2011baby}
Kulkarni, G.; Premraj, V.; Dhar, S.; Li, S.; Choi, Y.; Berg, A.~C.; and Berg,
  T.~L.
\newblock 2011.
\newblock Baby talk: Understanding and generating simple image descriptions.
\newblock  1601--1608.

\bibitem[\protect\citeauthoryear{Liu \bgroup et al\mbox.\egroup
  }{2018}]{liu2018}
Liu, F.; Ren, X.; Liu, Y.; Wang, H.; and Sun, X.
\newblock 2018.
\newblock Stepwise image-topic merging network for generating detailed and
  comprehensive image captions.
\newblock In {\em EMNLP 2018}.

\bibitem[\protect\citeauthoryear{Ma \bgroup et al\mbox.\egroup
  }{2018}]{DBLP:journals/corr/abs-1803-01465}
Ma, S.; Sun, X.; Li, W.; Li, S.; Li, W.; and Ren, X.
\newblock 2018.
\newblock Query and output: Generating words by querying distributed word
  representations for paraphrase generation.
\newblock In {\em {NAACL-HLT} 2018},  196--206.

\bibitem[\protect\citeauthoryear{Mao \bgroup et al\mbox.\egroup
  }{2014}]{mao2014explain}
Mao, J.; Xu, W.; Yang, Y.; Wang, J.; and Yuille, A.~L.
\newblock 2014.
\newblock Explain images with multimodal recurrent neural networks.
\newblock {\em arXiv: Computer Vision and Pattern Recognition}.

\bibitem[\protect\citeauthoryear{Pasunuru and Bansal}{2017}]{pasunuru2017multi}
Pasunuru, R., and Bansal, M.
\newblock 2017.
\newblock Multi-task video captioning with video and entailment generation.
\newblock {\em arXiv preprint arXiv:1704.07489}.

\bibitem[\protect\citeauthoryear{Ramanishka \bgroup et al\mbox.\egroup
  }{2016}]{ramanishka2016multimodal}
Ramanishka, V.; Das, A.; Park, D.~H.; Venugopalan, S.; Hendricks, L.~A.;
  Rohrbach, M.; and Saenko, K.
\newblock 2016.
\newblock Multimodal video description.
\newblock In {\em Proceedings of the 2016 ACM on Multimedia Conference},
  1092--1096.
\newblock ACM.

\bibitem[\protect\citeauthoryear{Rohrbach \bgroup et al\mbox.\egroup
  }{2014}]{tacosml}
Rohrbach, A.; Rohrbach, M.; Qiu, W.; Friedrich, A.; Pinkal, M.; and Schiele, B.
\newblock 2014.
\newblock Coherent multi-sentence video description with variable level of
  detail.
\newblock In {\em {GCPR} 2014},  184--195.

\bibitem[\protect\citeauthoryear{Rohrbach \bgroup et al\mbox.\egroup
  }{2015}]{vd1}
Rohrbach, A.; Rohrbach, M.; Tandon, N.; and Schiele, B.
\newblock 2015.
\newblock A dataset for movie description.
\newblock In {\em {CVPR} 2015},  3202--3212.

\bibitem[\protect\citeauthoryear{Shen \bgroup et al\mbox.\egroup
  }{2017}]{shen2017weakly}
Shen, Z.; Li, J.; Su, Z.; Li, M.; Chen, Y.; Jiang, Y.-G.; and Xue, X.
\newblock 2017.
\newblock Weakly supervised dense video captioning.
\newblock In {\em CVPR 2017}, volume~2, ~10.

\bibitem[\protect\citeauthoryear{Shetty and Laaksonen}{2016}]{shetty2016frame}
Shetty, R., and Laaksonen, J.
\newblock 2016.
\newblock Frame-and segment-level features and candidate pool evaluation for
  video caption generation.
\newblock In {\em Proceedings of the 2016 ACM on Multimedia Conference},
  1073--1076.
\newblock ACM.

\bibitem[\protect\citeauthoryear{Sutskever, Vinyals, and
  Le}{2014}]{sutskever2014sequence}
Sutskever, I.; Vinyals, O.; and Le, Q.~V.
\newblock 2014.
\newblock Sequence to sequence learning with neural networks.
\newblock In {\em Advances in neural information processing systems},
  3104--3112.

\bibitem[\protect\citeauthoryear{Tapaswi \bgroup et al\mbox.\egroup
  }{2016}]{movieqa}
Tapaswi, M.; Zhu, Y.; Stiefelhagen, R.; Torralba, A.; Urtasun, R.; and Fidler,
  S.
\newblock 2016.
\newblock Movieqa: Understanding stories in movies through question-answering.
\newblock In {\em {CVPR} 2016},  4631--4640.

\bibitem[\protect\citeauthoryear{Torabi \bgroup et al\mbox.\egroup
  }{2015}]{mvad}
Torabi, A.; Pal, C.~J.; Larochelle, H.; and Courville, A.~C.
\newblock 2015.
\newblock Using descriptive video services to create a large data source for
  video annotation research.
\newblock {\em CoRR} abs/1503.01070.

\bibitem[\protect\citeauthoryear{Vaswani \bgroup et al\mbox.\egroup
  }{2017}]{transformer}
Vaswani, A.; Shazeer, N.; Parmar, N.; Uszkoreit, J.; Jones, L.; Gomez, A.~N.;
  Kaiser, L.; and Polosukhin, I.
\newblock 2017.
\newblock Attention is all you need.
\newblock In {\em NIPS 2017},  6000--6010.

\bibitem[\protect\citeauthoryear{Venugopalan \bgroup et al\mbox.\egroup
  }{2015a}]{vd2}
Venugopalan, S.; Rohrbach, M.; Donahue, J.; Mooney, R.~J.; Darrell, T.; and
  Saenko, K.
\newblock 2015a.
\newblock Sequence to sequence - video to text.
\newblock In {\em {ICCV} 2015},  4534--4542.

\bibitem[\protect\citeauthoryear{Venugopalan \bgroup et al\mbox.\egroup
  }{2015b}]{vd3}
Venugopalan, S.; Xu, H.; Donahue, J.; Rohrbach, M.; Mooney, R.~J.; and Saenko,
  K.
\newblock 2015b.
\newblock Translating videos to natural language using deep recurrent neural
  networks.
\newblock In {\em {NAACL} {HLT} 2015},  1494--1504.

\bibitem[\protect\citeauthoryear{Vinyals \bgroup et al\mbox.\egroup
  }{2015}]{vinyals2015show}
Vinyals, O.; Toshev, A.; Bengio, S.; and Erhan, D.
\newblock 2015.
\newblock Show and tell: A neural image caption generator.
\newblock In {\em CVPR 2015},  3156--3164.

\bibitem[\protect\citeauthoryear{Xu \bgroup et al\mbox.\egroup
  }{2015}]{xu2015show}
Xu, K.; Ba, J.; Kiros, R.; Cho, K.; Courville, A.; Salakhudinov, R.; Zemel, R.;
  and Bengio, Y.
\newblock 2015.
\newblock Show, attend and tell: Neural image caption generation with visual
  attention.
\newblock In {\em ICML 2015},  2048--2057.

\bibitem[\protect\citeauthoryear{Xu \bgroup et al\mbox.\egroup }{2016}]{msrvtt}
Xu, J.; Mei, T.; Yao, T.; and Rui, Y.
\newblock 2016.
\newblock {MSR-VTT:} {A} large video description dataset for bridging video and
  language.
\newblock In {\em {CVPR} 2016},  5288--5296.

\bibitem[\protect\citeauthoryear{Xu \bgroup et al\mbox.\egroup
  }{2018}]{DBLP:journals/corr/abs-1805-05181}
Xu, J.; Sun, X.; Zeng, Q.; Ren, X.; Zhang, X.; Wang, H.; and Li, W.
\newblock 2018.
\newblock Unpaired sentiment-to-sentiment translation: {A} cycled reinforcement
  learning approach.
\newblock In {\em {ACL} 2018},  979--988.

\end{thebibliography}
\end{document}